\documentclass[11pt]{article}

\usepackage[final]{acl}

\usepackage{times}
\usepackage{latexsym}
\usepackage{amsmath}
\usepackage{amssymb}

\usepackage{marvosym}

\usepackage[T1]{fontenc}

\usepackage[utf8]{inputenc}

\usepackage{microtype}

\usepackage{inconsolata}

\usepackage{graphicx}
\usepackage{booktabs}
\usepackage{float}

\newcommand{\tabhead}[1]{\shortstack[c]{#1}}
\providecommand{\authoref}[1]{\autoref{#1}}

\title{VLA Grounder: Language-Conditioning Space Optimization for Black-Box VLA Models}

\author{
 \textbf{Damir Shodiev\textsuperscript{2}},
 \textbf{Nikita Kachaev\textsuperscript{1,4}},
 \textbf{Alexey K. Kovalev\textsuperscript{1,2\Letter}},
 \textbf{Aleksandr I. Panov\textsuperscript{1,2}},
\\
 \textbf{Aleksei Staroverov\textsuperscript{1,2,3}},
\\
\\
 \textsuperscript{1}AXXX,
 \textsuperscript{2}MIRAI,
 \textsuperscript{3}MISIS,
 \textsuperscript{4}ITMO University AI Talent Hub,
\\
 \small{
   \Letter\ \textbf{Correspondence:} \href{mailto:kovalev.a@miriai.org}{kovalev.a@miriai.org}
 }
}

\begin{document}

\maketitle
\begin{abstract}
  Vision-Language-Action (VLA) models are commonly treated as end-to-end action policies conditioned on natural-language task descriptions. In practice, however, their behavior often depends sharply on how the instruction is phrased, suggesting that language is not merely a task label but an optimizable conditioning input. We study whether frozen VLA policies can be improved by optimizing language space rather than updating action weights. Our method introduces a language-conditioning space policy that translates a human instruction into a short VLA-grounded command using object appearance, spatial relations, and target-grounding cues. The language-conditioning space policy is optimized with reinforcement learning from sparse task-completion rewards, while the downstream VLA remains fully frozen. Experiments on RL4VLA and VL-Think show that language-conditioning space optimization improves success on instruction-sensitive, symbolic, and multi-object manipulation tasks, demonstrating that language can serve as an optimizable variable for robot foundation models.

\textbf{Website}: \url{https://cognitiveaisystems.github.io/VLA-Grounder/}

\end{abstract}

\section{Introduction}

\begin{figure}[t]
  \centering
  \includegraphics[width=\linewidth]{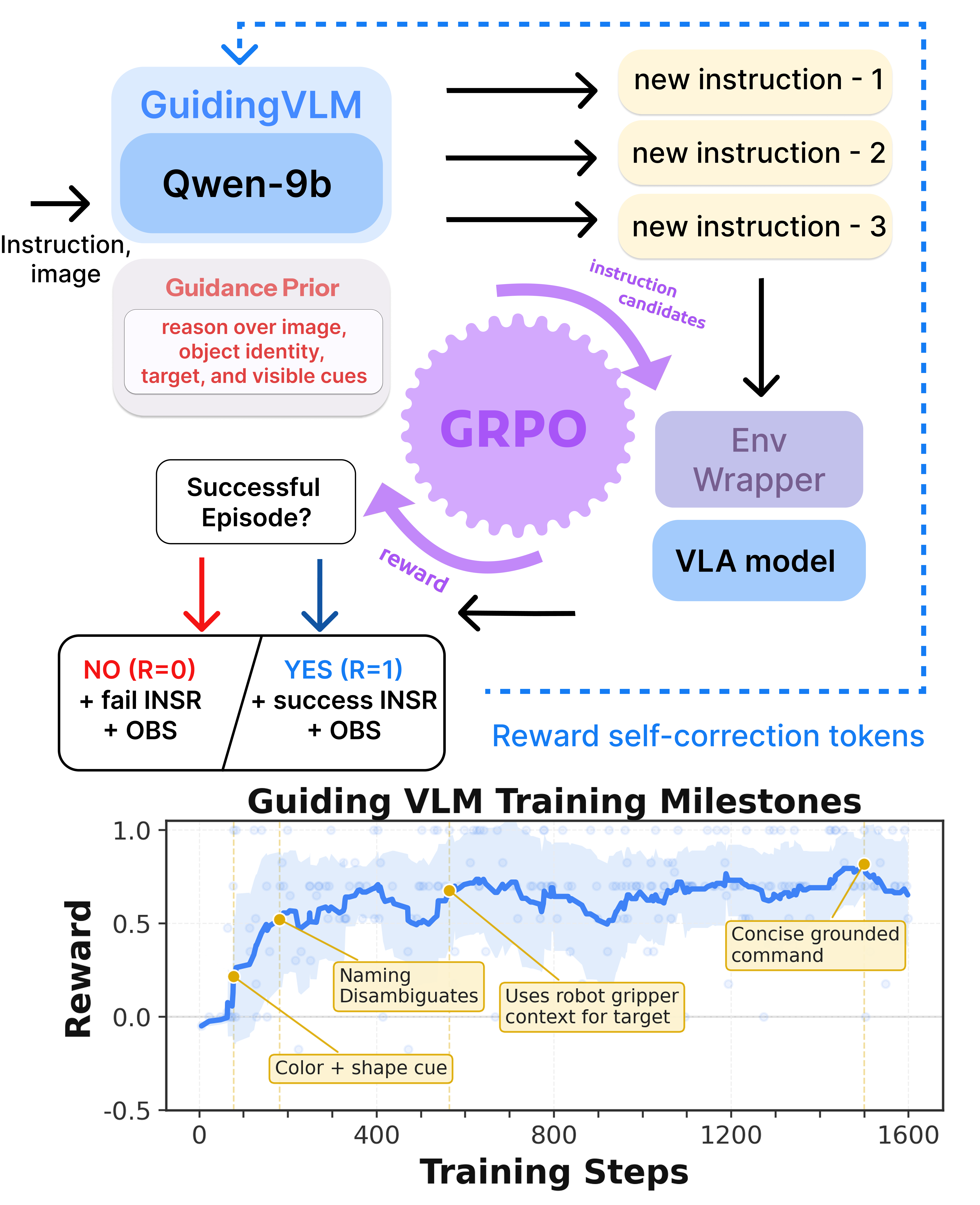}
  \caption{Visual overview of language-conditioning space optimization. A failure-derived command-space prior conditions a pre-trained VLM to translate the user instruction and scene image into a short VLA-grounded command. This command conditions a VLA policy, improving action success without updating VLA weights.}
  \label{fig:visual_abstract}
\end{figure}

Vision-Language-Action (VLA) models represent a promising paradigm in robotic learning, combining visual perception, natural language understanding, and action generation within a single policy. Recent VLA models leverage large-scale pre-training and robotic demonstrations to produce actions directly from images and language, enabling broad zero-shot manipulation across tasks and embodiments~\cite{black2024pi0,zhang2025experiences,zhong2025survey}.

Despite this progress, the language input to a VLA is often treated as a passive task description, even though it directly shapes the action distribution of the policy. In practice, VLA behavior is highly sensitive to how a goal is expressed~~\cite{moskalenko2025apple, blindvla, fang2025intention}. A command such as \textit{"put bread on plate"} may fail even when the scene contains a visually clear target, while a more grounded command such as \textit{"pick up the brown round object and put it on the yellow plate"} can succeed. Similar effects appear when an object name is visually misleading: \textit{"put champagne glass on plate"} can be improved by referring to the same object as \textit{"the tall white glass"}. These examples suggest that many VLA failures arise from an instruction-to-grounding mismatch between human semantic intent and the perceptual categories available to the robot policy.

This mismatch is amplified by the limited linguistic diversity of robot demonstration datasets~\cite{walke2024bridgedatav2datasetrobot, embodimentcollaboration2025openxembodimentroboticlearning}. Most datasets contain only short, templated instructions and everyday household objects, while user requests may include ambiguous objects, abstract references, or uncommon categories. As a result, the language channel of a VLA is underused: models reserve substantial capacity for text, but the instruction often fails to provide the visual and spatial cues needed for robust action selection~\cite{zhang2025experiences,liu2025what,black2024pi0,staroverov2025semantic}.

A natural response is to modify the instruction rather than the VLA weights. We therefore propose to insert an additional language-conditioning space policy between the human instruction and the frozen VLA, and ask whether command selection can provide an effective adaptation mechanism for large robot foundation models without updating their action weights.

We instantiate this idea with a language-conditioning space policy placed upstream of a frozen VLA, as shown in \authoref{fig:visual_abstract}. Given the current image and the original human instruction, the language-conditioning space policy produces a short VLA-grounded command that exposes execution-relevant cues: object appearance, color, shape, spatial position, relative location, and simplified target descriptions. The generated command is then used as the language input to the unchanged VLA policy. Because the downstream action model is kept fixed, the method adapts the conditioning signal rather than the action policy, making it non-invasive for different frozen VLA models.

We train the language-conditioning space policy (VLA Grounder) with Reinforcement Learning (RL) from sparse task-completion rewards while keeping the downstream VLA fully frozen (\authoref{fig:visual_abstract}). This casts adaptation as \emph{language-conditioning space optimization}: the learned decision variable is not a continuous robot action or a VLA weight update, but the VLA-grounded command through which the frozen action model receives and grounds the task. To make this optimization meaningful, we use a failure-derived command-space prior that restricts exploration to concise, visually grounded, and VLA-compatible commands. Within this structured command space, RL provides an action-grounded signal for discovering which commands most reliably elicit successful behavior from the existing policy. Experiments on VL-Think~\cite{blindvla} and RL4VLA~\cite{liu2025what} benchmarks show that our method creates better grounded guidance commands for frozen $\pi_0$ and OpenVLA models, increasing downstream task success across object grounding and multi-object manipulation settings.

The contributions of this work are as follows:

\begin{enumerate}
    \item \textbf{Language-conditioning space optimization for VLA models}: We formulate VLA adaptation as learning a language-conditioning space policy that maps human instructions to VLA-grounded commands while keeping the downstream action policy fixed.

    \item \textbf{RL in language-conditioning space}: We introduce VLA Grounder, language-conditioning space policy that maps an image and human instruction to a concise VLA-grounded command. We optimize only the language-conditioning space policy from sparse task-completion rewards, showing that downstream robot behavior can improve without updating VLA action weights.

    \item \textbf{Empirical and diagnostic evidence}: We evaluate on VL-Think and RL4VLA with frozen $\pi_0$ and OpenVLA backbones. The results show improved success on symbolic grounding, multi-object manipulation, and fixed-object settings, and our analyses indicate that optimized commands provide more grounded conditioning signals for the downstream VLA.

\end{enumerate}

\section{Related Work}

\subsection{Language-Conditioned VLA Policies}

Large-scale robot policies increasingly separate high-level semantic control from low-level motor generation. ACT introduced action chunking for manipulation~\cite{zhao2023aloha}, RT-2 transferred web-scale vision-language pretraining into robot action generation~\cite{zitkovich2023rt2}, and recent generalist policies such as OpenVLA and $\pi_0$ scale this recipe to broad zero-shot manipulation~\cite{kim2024openvla,black2024pi0}. Embodied chain-of-thought and $\pi_{0.5}$ further push toward richer intermediate reasoning and semantic control over action generation~\cite{zawalski2024ecot,black2025pi05}. Another line of work makes a VLA more controllable by enriching its conditioning signal. Higher-level semantic prediction in $\pi_{0.5}$, semantic guidance methods for VLA robustness, and prompt optimization methods in vision-language settings all support the broader idea that behavior depends strongly on how context is represented~\cite{black2025pi05,zhan2026stable,lee2023readonly}. These approaches treat language, metadata, or intermediate semantic variables as mechanisms for conditioning behavior.

Our paper asks a complementary post-training question: given an already frozen VLA, can we learn a better VLA-grounded command for control?

\subsection{RL Adaptation of VLA Models}

Reinforcement learning has become a central tool for improving VLA policies after imitation learning. RL4VLA studies PPO~\cite{schulman2017proximalpolicyoptimizationalgorithms}-, GRPO~\cite{shao2024deepseekmathpushinglimitsmathematical}-, and preference-based fine-tuning for VLA generalization~\cite{liu2025what}; $\pi_{\texttt{RL}}$ extends online RL to flow-based VLA models~\cite{chen2025pirl}; and $\pi^*_{0.6}$~\cite{physicalintelligence2025pistar} shows how deployment and corrective data can substantially improve a VLA through continued learning. More broadly, reward optimization for language policies, including ArCHer, PReWrite, and StablePrompt, shows that sparse reward can shape text-producing policies in useful ways~\cite{zhou2024archer,kong2024prewrite,kwon2024stableprompt}.

In contrast, our method adapts the language policy itself, not the VLA model weights, enabling parameter-efficient, non-invasive language-side adaptation.

\subsection{VLA Language Robustness}

Recent work shows that language remains an underdeveloped modality in embodied AI. Dataset audits find highly repetitive and template-like instructions in robot corpora~\cite{wanna2026linguistic}. Stable Language Guidance frames related failures as modality imbalance, where strong visual priors overpower sparse linguistic signals~\cite{zhan2026stable}, adversarial paraphrases and irrelevant context can significantly degrade VLA behavior even when task intent is preserved~\cite{moskalenko2025apple,spiridonov2025sparta}, and mechanistic analyses suggest that language-conditioned behavior is present internally but not always reliably exposed by the input instruction~\cite{haon2025mechanistic}. We build on this diagnosis but reframe it as an optimization problem over command space. Instead of treating language brittleness as only a robustness issue, we treat language conditioning as an actionable input space that can be optimized for downstream execution success.

\begin{figure*}[ht]
\begin{center}
\centerline{\includegraphics[width=1.8\columnwidth]{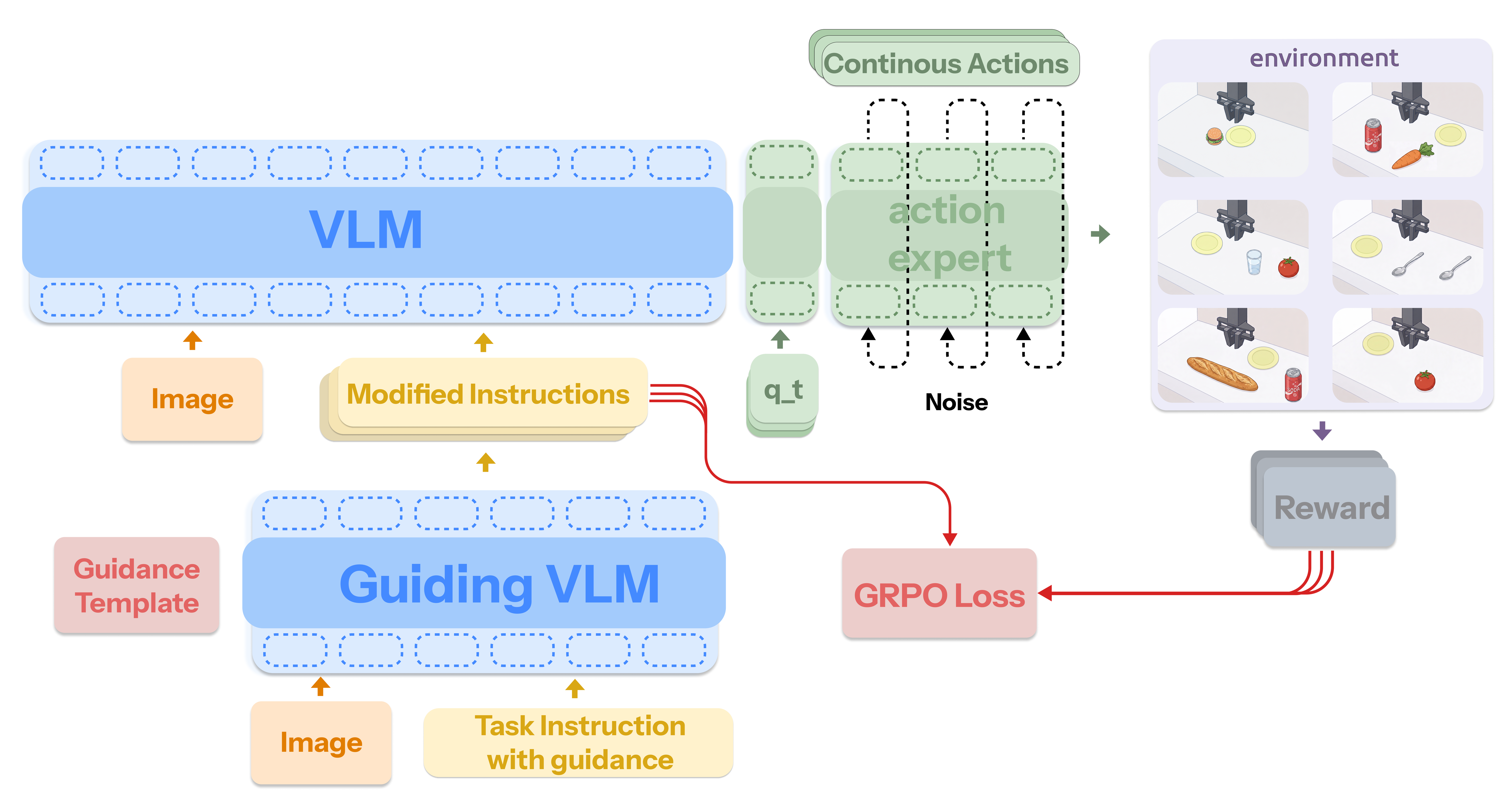}}
\caption{Architecture and training. A language-conditioning space policy maps the task instruction, scene image, and a failure-derived command-space prior to a VLA-grounded command, which is executed by a frozen VLA. Only this policy is trained with RL from sparse task-completion rewards; VLA weights remain frozen.}
\label{fig:model_scheme}
\end{center}
\end{figure*}

\section{Problem Formulation}

\subsection{Partially Observable Control with Language Instructions}

We consider an episodic partially observable Markov decision process (POMDP) $\mathcal{M} = (\mathcal{S}, \mathcal{A}, \mathcal{O}, P, Z, r, \rho_0, \gamma)$, where $\mathcal{S}$ is the latent state space, $\mathcal{A}$ the primitive action space, $\mathcal{O}$ the observation space, $P$ the transition kernel, $Z$ the observation kernel, $r$ the reward function, $\rho_0$ the initial-state distribution, and $\gamma \in [0,1)$ the discount factor. At time $t$, the environment occupies state $s_t$, emits $o_t \sim Z(\cdot \mid s_t)$, receives action $a_t$, produces reward $r(s_t,a_t)$, and transitions to $s_{t+1} \sim P(\cdot \mid s_t,a_t)$. Each episode is paired with a fixed human instruction $u \in \mathcal{U}_{\mathrm{H}}$. We write the discounted return of a trajectory $\tau$ as $R(\tau) = \sum_{t=0}^{T-1} \gamma^t r_t$.

\subsection{Frozen Vision-Language-Action Policies}

A pretrained Vision-Language-Action policy maps an observation and a language command to an action distribution, $\pi_{\mathrm{VLA}} : \mathcal{O} \times \mathcal{U} \to \Delta(\mathcal{A})$, where $\mathcal{U}$ is the space of commands consumed by the action policy. In standard deployment, the raw instruction is passed directly to the policy, so $a_t \sim \pi_{\mathrm{VLA}}(\cdot \mid o_t, u)$. More generally, each command $\tilde{u} \in \mathcal{U}$ induces a frozen low-level policy $\pi_{\mathrm{VLA}}^{\tilde{u}}(a \mid o) := \pi_{\mathrm{VLA}}(a \mid o,\tilde{u})$. Distinct commands that preserve the same task intent may still induce different grounded behavior.

Let $\mathcal{U}(u) \subseteq \mathcal{U}$ denote the command space associated with instruction $u$, i.e., the set of commands that are semantically consistent with $u$ but may differ in wording, specificity, or grounding cues. The goal is not to learn a new low-level action policy, but to exploit this command space to obtain better behavior from a frozen one.

\subsection{Problem Setting}

We are given a pretrained frozen VLA policy $\pi_{\mathrm{VLA}}$ and seek to maximize return in $\mathcal{M}$ by selecting a command $\tilde{u} \in \mathcal{U}(u)$ rather than by changing action-model weights. Equivalently, the learned component chooses which member of the frozen family $\{\pi_{\mathrm{VLA}}^{\tilde{u}} : \tilde{u} \in \mathcal{U}(u)\}$ will generate actions during the rollout. This formulation is useful only when the policy is sensitive to language-space variation, meaning that semantically valid commands can induce meaningfully different behavior. We assume only black-box access: for any observation $o$ and command $\tilde{u}$, we may sample from $\pi_{\mathrm{VLA}}(\cdot \mid o,\tilde{u})$, but we do not use weights, gradients, or internal activations.

\section{Method}
\subsection{RL in Language-Conditioning space}
In standard deployment, a VLA acts under the raw instruction $u$. We instead treat the command itself as the high-level decision variable: at observation $o$, we first select a command $\tilde{u} \in \mathcal{U}(u)$ and then sample the primitive action from $\pi_{\mathrm{VLA}}(\cdot \mid o,\tilde{u})$. This induces an action-space transformation from primitive actions to language commands.

\begin{equation}
\begin{aligned}
P^{\mathrm{lang}}(s' \mid s,\tilde{u})
&:= \mathbb{E}_{\substack{o \sim Z(\cdot \mid s) \\
    a \sim \pi_{\mathrm{VLA}}^{\tilde{u}}(\cdot \mid o)}}
    \bigl[P(s' \mid s,a)\bigr], \\
r^{\mathrm{lang}}(s,\tilde{u})
&:= \mathbb{E}_{\substack{o \sim Z(\cdot \mid s) \\
    a \sim \pi_{\mathrm{VLA}}^{\tilde{u}}(\cdot \mid o)}}
    \bigl[r(s,a)\bigr].
\end{aligned}
\label{eq:lang-mdp-kernels}
\end{equation}
These induced kernels define a transformed control problem, denoted $\mathcal{M}^{\mathrm{lang}}_u$, whose action space is $\mathcal{U}(u)$ and whose dynamics and rewards are given by $P^{\mathrm{lang}}$ and $r^{\mathrm{lang}}$. This is the sense in which the adaptation problem can be cast as reinforcement learning in language space. We focus on rollout-level optimization: a policy $q_{\phi}(\tilde{u} \mid o_0, u)$ is invoked once at the beginning of the episode, outputs a single command, and keeps it fixed for the full rollout. Sampling $\tilde{u} \sim q_{\phi}(\cdot \mid o_0,u)$ therefore selects one member of the frozen family $\{\pi_{\mathrm{VLA}}^{\tilde{u}} : \tilde{u} \in \mathcal{U}(u)\}$.

\begin{equation}
\begin{aligned}
J(\phi)
&=
\mathbb{E}_{(o_0,u)\sim\mathcal{D}}
\mathbb{E}_{\tilde{u}\sim q_{\phi}(\cdot \mid o_0,u)} \\
&\qquad
\mathbb{E}_{\tau \sim p(\cdot \mid o_0,\pi_{\mathrm{VLA}}^{\tilde{u}})}
\bigl[R(\tau)\bigr].
\end{aligned}
\label{eq:lang-objective}
\end{equation}
where $\mathcal{D}$ is the distribution of initial image-instruction pairs. The raw frozen-VLA baseline is recovered by the degenerate policy that always returns the original instruction.

\subsection{Language Aliasing and Structured Exploration}

Language-Conditioning space has additional structure that naive RL does not exploit. In particular, there may exist distinct commands $\tilde{u} \neq \tilde{u}'$ such that $\pi_{\mathrm{VLA}}(\cdot \mid o,\tilde{u}) \approx \pi_{\mathrm{VLA}}(\cdot \mid o,\tilde{u}')$ for many observations $o$; we refer to this as \emph{language aliasing}. It is common for shallow rewrites, where grammatical edits or near-synonymous substitutions are lexically different but behaviorally redundant. Our method therefore biases exploration toward short executable commands that vary along grounding-relevant dimensions such as source-object description, target cues, color, shape, and spatial location.

\subsection{Optimizing the Guiding VLM Policy}

We instantiate $q_{\phi}$ with a Guiding VLM and train it with GRPO~\cite{shao2024deepseekmathpushinglimitsmathematical} using only scalar rollout reward. \authoref{fig:model_scheme} summarizes the language-conditioning space optimization loop: the policy receives the initial image and instruction, proposes candidate commands, and the frozen VLA policy executes a rollout under each one. For each training pair $(o_i,u_i)$, we sample a group of candidate commands $\{\tilde{u}_i^{(k)}\}_{k=1}^{G}$ from $q_{\phi}(\cdot \mid o_i,u_i)$, execute the frozen VLA under each command, and compare the resulting returns $\{R_i^{(k)}\}_{k=1}^{G}$ within the group. We use the normalized advantage
\begin{equation}
\hat{A}_i^{(k)}
=
\frac{R_i^{(k)}-\bar{R}_i}{\sigma_i+\varepsilon},
\qquad
\bar{R}_i
=
\frac{1}{G}\sum_{k=1}^{G} R_i^{(k)},
\label{eq:group-advantage}
\end{equation}
where $\sigma_i$ is the empirical standard deviation of rewards in the group. Commands with larger relative advantage are up-weighted by the next update, while weaker ones are down-weighted. \autoref{fig:algorithm} summarizes the procedure.

\subsection{VLA Black-Box Deployment}

At inference time, the system receives a new image and instruction, samples one command $\hat{\tilde{u}}$ from the trained language-conditioning space policy, and conditions the frozen VLA policy on that command for the full rollout. There is no reward query, no RL update, and no action-model modification at test time. VLA Grounder changes only the language-conditioning input and requires only black-box forward access to the downstream action policy. In this paper, we evaluate this setup on frozen $\pi_0$ and OpenVLA backbones.

\subsection{Why Language Exploration Matters}

Group-based RL algorithms, such as GRPO, derive a signal by comparing multiple samples for the same task instance. If the sampled set differs only in grammar, then the reward cannot meaningfully separate better and worse conditioning signals. By contrast, if samples vary in robot-relevant grounding cues, for example, color, left-right position, object description, or target relation---the optimization process can assign credit to language transformations that actually change downstream behavior.

\begin{table*}[t]
\centering
\small
\setlength{\tabcolsep}{3pt}
\renewcommand{\arraystretch}{1.05}
\begin{tabular*}{\textwidth}{@{\extracolsep{\fill}}lcccccc}
\toprule
& \multicolumn{2}{c}{$\pi_0$}
& \multicolumn{4}{c}{OpenVLA} \\
\cmidrule(lr){2-3}
\cmidrule(lr){4-7}
Task
& orig
& Qwen3.5-9B
& orig
& Qwen3.5-9B
& TextGrad
& GEPA \\
\midrule
Arrow
& $6.2 \pm 1.3$
& $\mathbf{36.5 \pm 3.2}$
& $13.5 \pm 3.2$
& $\mathbf{62.5 \pm 1.3}$
& $44.0 \pm 2.5$
& $51.3 \pm 2.8$ \\

Color
& $\mathbf{45.3 \pm 7.1}$
& $44.8 \pm 7.0$
& $55.7 \pm 8.7$
& $\mathbf{82.8 \pm 2.6}$
& $61.2 \pm 3.6$
& $73.2 \pm 1.2$ \\

Laundry
& $16.1 \pm 0.7$
& $\mathbf{26.6 \pm 2.6}$
& $15.6 \pm 2.6$
& $37.5 \pm 3.8$
& $19.1 \pm 3.6$
& $\mathbf{47.0 \pm 2.1}$ \\

Public Info
& $14.6 \pm 1.5$
& $\mathbf{43.2 \pm 2.9}$
& $21.4 \pm 3.7$
& $\mathbf{65.6 \pm 3.4}$
& $25.6 \pm 1.6$
& $48.5 \pm 3.1$ \\

Shape
& $20.3 \pm 5.8$
& $\mathbf{49.5 \pm 6.0}$
& $27.6 \pm 1.5$
& $\mathbf{74.5 \pm 6.4}$
& $54.2 \pm 3.3$
& $59.5 \pm 4.1$ \\

Traffic
& $6.2 \pm 1.3$
& $\mathbf{37.0 \pm 10.2}$
& $16.7 \pm 2.7$
& $\mathbf{63.0 \pm 4.8}$
& $35.8 \pm 3.4$
& $49.4 \pm 5.1$ \\

Weather
& $19.3 \pm 2.7$
& $\mathbf{42.2 \pm 3.4}$
& $20.3 \pm 4.6$
& $\mathbf{61.5 \pm 3.9}$
& $30.4 \pm 4.1$
& $57.2 \pm 3.4$ \\
\midrule
Avg.
& $18.3$
& $\mathbf{40.0}$
& $24.4$
& $\mathbf{63.9}$
& $38.6$
& $55.2$ \\
\bottomrule
\end{tabular*}
\caption{Main results on VL-Think. All VLA backbones are frozen. The language-conditioning space policy uses the strongest structured reasoning configuration, Qwen3.5-9B.}
\label{tab:vlthnik-main}
\end{table*}

\section{Experiments}

We evaluate on two benchmark families. VL-Think~~\cite{blindvla} isolates visual-language reasoning under fixed manipulation: the robot places the same source object on a target board matching an abstract concept such as an arrow, sign, shape or weather icon. RL4VLA~~\cite{liu2025what} tests broader semantic generalization; we use \textit{MultiPlate}, \textit{MultiCarrot}, and fixed-object \textit{Pepper}, which require distractor handling and object-target grounding. We also evaluate frozen MolmoAct2~~\cite{fang2026molmoact2} on a physical SO-101. All reported VLA backbones are frozen, and we report success rate as mean $\pm$ standard deviation.

\begin{figure}[t]
  \centering
  \includegraphics[width=0.99\linewidth]{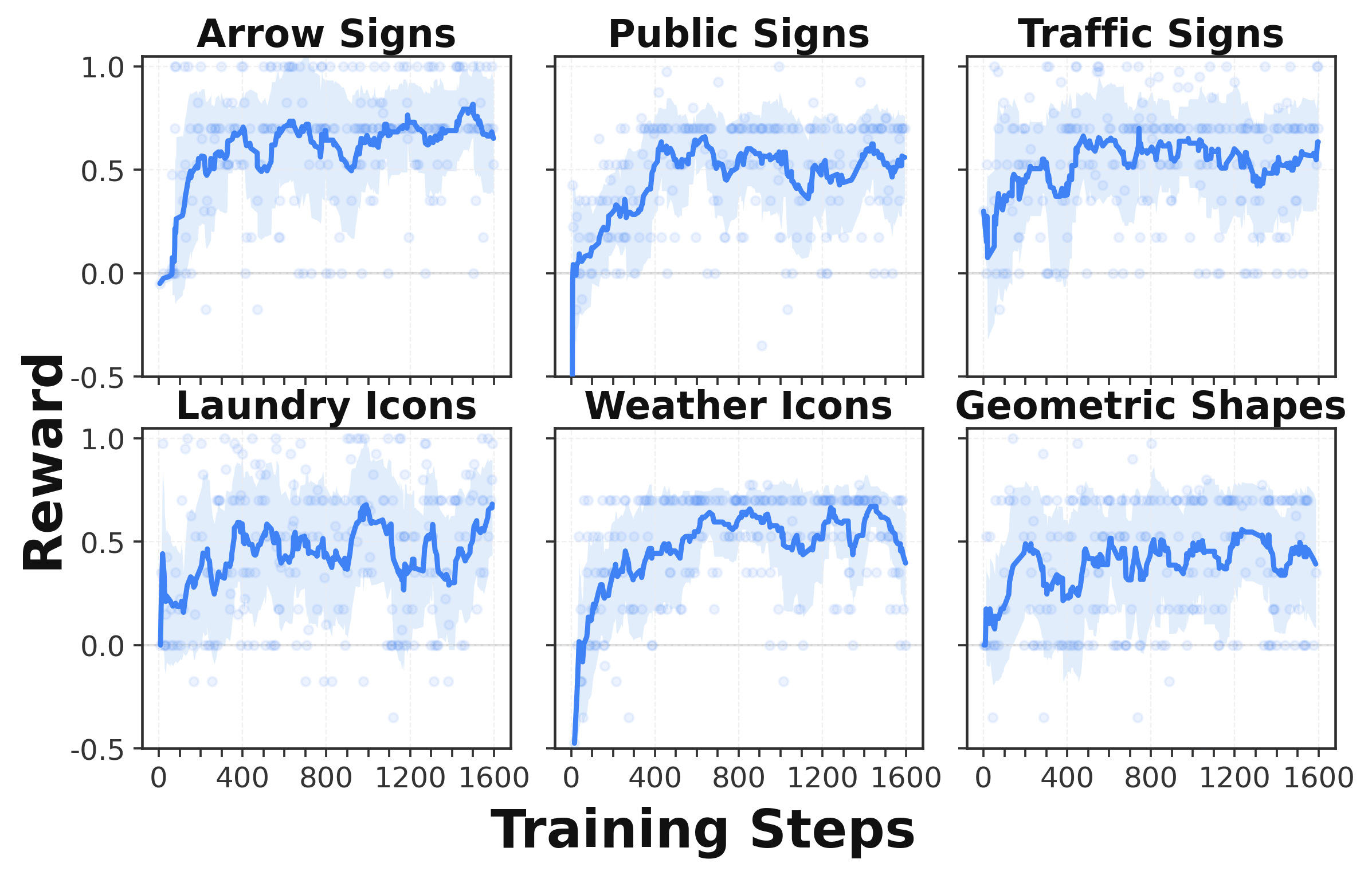}
  \caption{GRPO training for VL-Think tasks. Each panel reports rollout success rate over training steps for a symbolic grounding category. The upward trends indicate that reward updates to the language-conditioning space policy improve command selection while the downstream VLA action policy remains frozen.}
  \label{fig:language_aliasing}
\end{figure}

The primary baseline is the frozen VLA conditioned on the raw human instruction. We also compare against the structured language-conditioning space policy before RL (``w/o GRPO''), generic command rewriting without structured reasoning (``no reasoning''), GRPO without structured guidance (``GRPO w/o guidance''), and structured-guidance variants with different language-conditioning space policy sizes. To evaluate cross-backbone transfer, we additionally swap trained Grounder checkpoints between $\pi_0$ and OpenVLA and evaluate them with the other frozen VLA without retraining. For Pepper and VL-Think, we include TextGrad~~\cite{textgrad} and GEPA~~\cite{gepa}, two general-purpose prompt-optimization baselines that rewrite or mutate commands in language space. We organize the empirical section around main research questions:

\paragraph{RQ1: Does language-conditioning space optimization improve frozen VLA policies across backbones?}
Our experiments on VL-Think and RL4VLA (\autoref{tab:vlthnik-main}, \autoref{tab:rl4vla-main}, \autoref{fig:language_aliasing}) reveal a consistent effect: changing only the language input can substantially improve frozen action policies. The effect appears for both $\pi_0$ and OpenVLA, which suggests that the method is not exploiting a single backbone-specific quirk. The gains are most visible on VL-Think, where success depends heavily on converting abstract or symbolic targets into visually grounded commands. RL4VLA shows a smaller but still consistent trend, which is expected because these tasks add multi-object distractors and execution errors that cannot be removed by language alone.

\begin{figure}[b]
    \centering
    \includegraphics[width=1\linewidth]{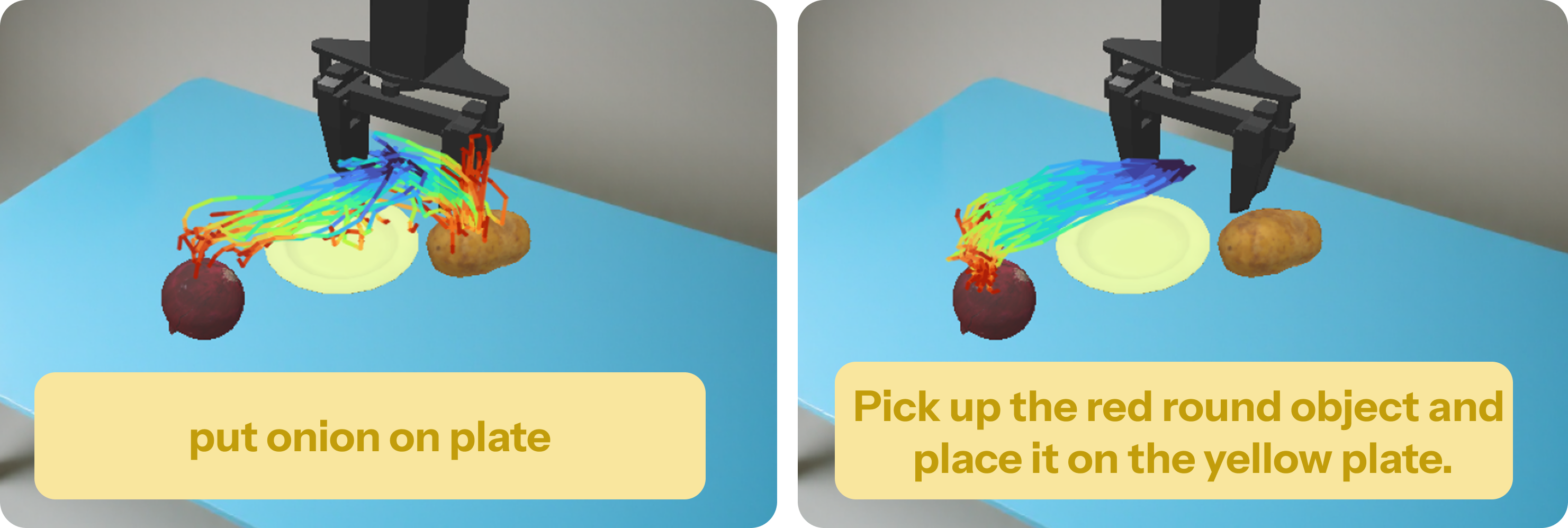}
    \caption{Trajectories comparison on RL4VLA MultiObject with frozen $\pi_0$. The default command under-specifies the object and causes unstable motion, while the rewritten command adds visible grounding cues and yields a more direct successful trajectory.}
    \label{fig:traces}
\end{figure}

\paragraph{RQ2: Does GRPO improve over the non-RL language-conditioning space policy and generic prompt-optimization baselines?}
The non-RL language-conditioning space policy already provides a useful command prior, but it does not reliably identify which command aliases the frozen action policy will execute. GRPO adds exactly this missing selection pressure: candidate commands are compared by downstream task reward rather than by linguistic plausibility. This distinction is clearest on VL-Think, where the trained policy separates sharply from its non-RL version. On RL4VLA the improvement is more modest, but remains positive despite higher action-side noise.

\paragraph{RQ3: How does reward-trained language optimization compare with traditional prompt optimization?}
We also tested generic prompt-optimization methods, including TextGrad and GEPA, using the same frozen-VLA setting. Our results in \authoref{tab:vlthnik-main} show that these methods can improve over the raw instruction baseline, but not nearly as much as GRPO-based RL adaptation. The reason is qualitative as well as quantitative: generic prompt optimizers often modify punctuation, wording, or writing style in ways that are hard to interpret as better robot grounding. VLA Grounder instead updates the language-conditioning space policy from rollout reward, so it favors command changes that actually alter execution success rather than commands that only look cleaner as text.

\begin{table*}[t]
\centering
\small
\setlength{\tabcolsep}{4pt}
\renewcommand{\arraystretch}{1.05}
\begin{tabular*}{\textwidth}{@{\extracolsep{\fill}}lcccc}
\toprule
& \multicolumn{2}{c}{$\pi_0$} & \multicolumn{2}{c}{OpenVLA} \\
\cmidrule(lr){2-3} \cmidrule(lr){4-5}
Task & orig & Qwen3.5-9B & orig & Qwen3.5-9B \\
\midrule
MultiPlate  & $16.7 \pm 7.0$ & $\mathbf{20.3 \pm 2.6}$ & $54.7 \pm 2.2$ & $\mathbf{64.6 \pm 5.2}$ \\
MultiCarrot & $24.5 \pm 0.7$ & $\mathbf{33.3 \pm 4.8}$ & $59.4 \pm 5.9$ & $\mathbf{63.5 \pm 5.3}$ \\
\midrule
Avg.        & $20.6$         & $\mathbf{26.8}$         & $57.0$         & $\mathbf{64.0}$ \\
\bottomrule
\end{tabular*}
\caption{Results on the RL4VLA semantic generalization tasks. The same structured language-conditioning space policy improves both frozen $\pi_0$ and frozen OpenVLA in multi-object and distractor-heavy settings.}
\label{tab:rl4vla-main}
\end{table*}

\begin{table*}[t]
\centering
\small
\setlength{\tabcolsep}{5pt}
\renewcommand{\arraystretch}{1.15}
\begin{tabular*}{\textwidth}{@{\extracolsep{\fill}}p{0.18\textwidth}p{0.24\textwidth}p{0.24\textwidth}p{0.24\textwidth}}
\toprule
Failure type & Raw instruction & Useful rewrite dimension & Language policy rewrite \\
\midrule
Abstract target & ``sunrise icon'' & visible symbol/color & ``white card with yellow sun'' \\
Rare object name & ``champagne glass'' & shape/material & ``tall white glass'' \\
Multi-object ambiguity & ``plate'' & spatial relation & ``left yellow plate'' \\
Weak object grounding & ``bread'' & color/shape & ``brown round object'' \\
\bottomrule
\end{tabular*}
\caption{Common instruction-grounding failures and useful command rewrite dimensions. The learned policy improves frozen VLA behavior by replacing abstract target names, rare object categories, ambiguous references, or weak object names with visually grounded descriptions that expose color, shape, material, symbol, or spatial cues.}
\label{tab:rewrite-dimensions}
\end{table*}

\begin{table}
\centering
\small
\setlength{\tabcolsep}{8pt}
\renewcommand{\arraystretch}{1.05}
\begin{tabular}{lc}
\toprule
Prompt & Accuracy \\
\midrule
Default & $66.6$ \\
VLA Grounder   & $89.3$ \\
\bottomrule
\end{tabular}
\caption{Linear probing accuracy on VL-Think Public Info tasks using OpenVLA hidden states under the default instruction and the rewritten command.}
\label{tab:rq4-probing}
\end{table}

\begin{table}[t]
\centering
\small

\setlength{\tabcolsep}{2pt}
\renewcommand{\arraystretch}{1.05}
\resizebox{\columnwidth}{!}{%
\begin{tabular}{@{}lcccc@{}}
\toprule
Task
& $\pi_0$ orig
& $\pi_0$ + TextGrad
& $\pi_0$ + GEPA
& $\pi_0$ + Qwen3.5 \\
\midrule
Pepper
& $39.1 \pm 6.6$
& $44.3 \pm 5.5$
& $61.7 \pm 4.1$
& $\mathbf{89.6 \pm 3.2}$ \\
\bottomrule
\end{tabular}%
}
\caption{Fixed-object RL4VLA result. On Pepper, the structured language-conditioning space policy yields a large gain with the $\pi_0$ action policy frozen.}
\label{tab:pepper}
\end{table}

\begin{table}[t]
\centering
\small
\begin{tabular}{p{0.54\columnwidth}p{0.32\columnwidth}}
\toprule
Audit statistic & Value \\
\midrule
Same source object and placement target
& $98.89\%$ $(6{,}345/6{,}416)$ \\

Mentions color
& $84.96\%$ \\

Mentions shape
& $18.87\%$ \\

Mentions both color and shape
& $17.44\%$ \\

Preserves the original object name
& $62.98\%$ \\
\bottomrule
\end{tabular}
\caption{Automatic scene-aware semantic audit of generated commands.}
\label{tab:semantic-audit}
\end{table}

\paragraph{RQ4: Do better commands lead to better VLA representations?}
We further probe whether improved commands merely change the final text input or also change the internal representations used by the VLA. Using VL-Think Public Info task, we compare OpenVLA representations under the original task instruction and under the rewritten command. The probing results \autoref{tab:rq4-probing}, \autoref{fig:traces} indicate that rewritten commands produce more informative prompt-conditioned representations: task-relevant distinctions become easier to recover from the VLA hidden states, and this aligns with the downstream success improvements. This suggests that language-conditioning space optimization changes the conditioning representation seen by the action model, rather than only changing a superficial instruction string.

\paragraph{RQ5: What command transformations does the learned policy discover?}
Based on prompt evolution analysis during training  (\authoref{tab:rewrite-dimensions}) we found that the learned policy tends to produce shorter and more executable commands rather than more verbose descriptions. On VL-Think, it often converts symbolic target labels into compact spatial or visual aliases, such as replacing an icon name with a reference to a white card at a particular position. On RL4VLA and Pepper, it usually keeps familiar receptacle names but rewrites brittle source-object names into visible color and shape descriptions. This pattern indicates that the policy is learning how to reduce the semantic burden placed on the frozen action model. The appendix provides \authoref{fig:prompt-example} prompt-format examples.

\paragraph{RQ6: Do rewrites preserve task semantics?}

Structured guidance defines the intended command space but is not a hard semantic constraint. We therefore audited all 6,416 saved MultiCarrot and MultiPlate commands across the train and test splits and both VLA backbones. For each command, we re-rendered the corresponding scene and used the base Qwen3.5-9B model to determine whether the rewrite retained the same source object and placement target.

As shown in \autoref{tab:semantic-audit}, 98.89\% of the commands ($6{,}345/6{,}416$) preserve both the source object and placement target. Color is mentioned in 84.96\% of the rewrites, shape in 18.87\%, and both color and shape in 17.44\%. The original object name is preserved in 62.98\% of the commands. Thus, only 1.11\% of the rewrites are judged to change the intended source or target, while most add visible grounding cues without replacing the task.

The lowest semantic-faithfulness result is 97.76\% on the MultiPlate test split. The observed failures include incorrect visual descriptions, confusion between watering cans and cups, and references to incorrect non-plate targets. Because the command generator and the automatic evaluator both belong to the Qwen3.5-9B model family, we report this result specifically as a large-scale scene-aware automatic audit rather than as human annotation.

\paragraph{RQ7: Does a simulation-trained Grounder transfer to a different VLA on a physical robot?}
Without additional adaptation, a Grounder trained with frozen $\pi_0$ in MultiCarrot simulation improves frozen MolmoAct2 on a physical SO-101 from $26/50$ to $38/50$ target-reaching successes. This transfer across simulator, backbone, and embodiment supports reusable grounding cues. \autoref{app:real-robot} gives the complete protocol and per-task results.

\section{Discussion}

\subsection{Command-Space Design, Reward Optimization, and Transfer}

The ablations suggest that the command space and the language model play different roles. Detailed per-task results for all ablations are reported in (\autoref{app:ablations-results}). Generic rewriting can already help, which confirms that the raw instruction is often a weak conditioning signal. Structured guidance improves the search space by pushing candidates toward object identity, visual attributes, and spatial grounding rather than shallow paraphrases. As shown in (\autoref{tab:guidance-grpo-ablation}), the matched ablations further separate the effects of structured guidance and reward optimization. The ``w/o GRPO'' configuration measures the benefit of the guidance prior alone, whereas ``GRPO w/o guidance'' tests reward-based optimization over a generic rewriting space. GRPO provides the larger individual improvement, while the full configuration achieves the strongest average performance, showing that structured exploration and reward-based selection are complementary. The model-scale results in (\autoref{tab:vlthnik-ablation}) further show that scaling the language model becomes more useful when the model has a richer space of possible aliases from which to choose. The benefit is not perfectly monotonic across every category, however, so scale alone does not explain the improvement; the strongest setting combines structured command variation with reward-based selection. The checkpoint-swap experiment (\autoref{tab:cross-backbone-transfer}) additionally shows that the learned Grounders are not restricted to the VLA backbone used during training: transferring checkpoints between $\pi_0$ and OpenVLA improves both frozen policies without retraining. This result indicates that the optimized commands capture reusable visual-language grounding cues in addition to backbone-specific linguistic preferences.

\subsection{Limits of Language-Conditioning Space Optimization}

The gains are largest when the user instruction is semantically valid but visually under-specified for the action policy. Symbolic VL-Think tasks such as arrows, public-information signs, weather icons, shapes, and traffic signs benefit because the learned policy can replace the symbolic target with a simpler visual or spatial description. Direct perceptual categories such as Color leave less room for improvement because the raw instruction already exposes a usable cue. The method is also limited on tasks where language is not the only bottleneck: clutter, distractors, wrong object grounding, severe visual misperception, missing low-level manipulation skill, or long-horizon reasoning demands can still dominate the rollout outcome.

\subsection{Broader Implication}

The broader implication of this work is that language can function as an optimizable adaptation layer for robot foundation models. If this view is correct, then future post-training of VLAs need not operate only in weight space or action space; it can also operate in the space of semantic conditioning signals that mediate between human intent and embodied action.

\section{Conclusion}

We study adaptation of frozen VLA policies through the language-conditioning input rather than through action-weight updates. The resulting perspective treats language as an optimizable conditioning variable and casts improvement as reinforcement learning in language space. VLA Grounder combines a language-conditioning space policy, a failure-derived command-space prior, and a sparse downstream task reward to discover VLA-grounded commands that better expose latent capability in frozen action policies. The central takeaway is simple but consequential: for large robot foundation models, policy improvement can be lifted from action space into language space.

\newpage

\section*{Limitations}

Our results show that language-conditioning space optimization can improve frozen VLA policies, but the scope of the evidence is still limited. The method also cannot repair all failures of a frozen action policy. Language-space optimization should be viewed as an adaptation layer over existing VLA capabilities rather than as a replacement for improving perception, robustness, control, or long-horizon reasoning.

\section*{Ethical Considerations}

This work studies language-conditioned robot control in benchmark settings and does not involve human-subject data.

\section*{Artifact Licenses}

We use publicly available benchmark and model artifacts, including VL-Think, RL4VLA, OpenVLA, INT-ACT, $\pi_0$, and Qwen, under their respective licenses and terms of use. Any released code or checkpoints will be distributed with an explicit license. Our use of these artifacts is limited to research evaluation.

\bibliography{custom}

\newpage
\appendix
\onecolumn

\section{Appendix Overview}

This appendix provides additional experimental results and implementation details supporting the main paper. \autoref{app:ablations-results} reports the model-scale, structured-guidance, GRPO, and cross-backbone-transfer ablations. \autoref{app:real-robot} presents the real-robot evaluation with a frozen MolmoAct2 policy on an SO-101 arm. \autoref{app:prompt-example} illustrates the prompt format, output contract, and model-matched prompt-optimization comparisons. \autoref{app:algorithm-loop} summarizes the GRPO training procedure and black-box optimization loop. Finally, \autoref{app:training-details} reports the language-policy configuration, optimization settings, rollout configuration, and computational cost of the reported training runs.

\section{Ablations}
\label{app:ablations-results}
\begin{table*}[h!]
\centering
\scriptsize
\setlength{\tabcolsep}{3.5pt}
\renewcommand{\arraystretch}{1.05}
\begin{tabular*}{\textwidth}{@{\extracolsep{\fill}}lccccc}
\toprule
Task & $\pi_0$ & \tabhead{+Qwen3-4B\\ no reasoning} & \tabhead{+Qwen3-4B\\reasoning} & \tabhead{+Qwen3.5-4B\\reasoning} & \tabhead{+Qwen3.5-9B\\reasoning} \\
\midrule
Arrow & $6.2 \pm 1.3$ & $11.5 \pm 7.1$ & $18.7 \pm 2.9$ & $17.3 \pm 3.0$ & $\mathbf{36.5 \pm 3.2}$ \\
Color & $\mathbf{45.3 \pm 7.1}$ & $28.8 \pm 6.8$ & $33.5 \pm 6.5$ & $35.0 \pm 6.0$ & $44.8 \pm 7.0$ \\
Laundry & $16.1 \pm 0.7$ & $21.4 \pm 2.5$ & $24.1 \pm 3.8$ & $25.7 \pm 2.3$ & $\mathbf{26.6 \pm 2.6}$ \\
Public Info & $14.6 \pm 1.5$ & $22.3 \pm 2.8$ & $20.9 \pm 3.1$ & $26.1 \pm 3.5$ & $\mathbf{43.2 \pm 2.9}$ \\
Shape & $20.3 \pm 5.8$ & $24.7 \pm 4.6$ & $33.4 \pm 5.8$ & $36.5 \pm 5.0$ & $\mathbf{49.5 \pm 6.0}$ \\
Traffic & $6.2 \pm 1.3$ & $23.5 \pm 3.2$ & $18.0 \pm 4.4$ & $21.1 \pm 3.8$ & $\mathbf{37.0 \pm 10.2}$ \\
Weather & $19.3 \pm 2.7$ & $25.4 \pm 3.9$ & $21.2 \pm 3.8$ & $23.9 \pm 4.1$ & $\mathbf{42.2 \pm 3.4}$ \\
\midrule
Avg. & $18.3$ & $22.5$ & $24.3$ & $26.5$ & $\mathbf{40.0}$ \\
\bottomrule
\end{tabular*}
\caption{Ablation on frozen $\pi_0$ over VL-Think benchmark. ``No reasoning'' denotes a generic command-rewriting policy; ``reasoning'' denotes the structured guidance used by our method.}
\label{tab:vlthnik-ablation}
\end{table*}

\begin{table*}[h!]
\centering
\scriptsize

\setlength{\tabcolsep}{2.5pt}
\renewcommand{\arraystretch}{1.08}
\begin{tabular*}{\textwidth}{@{\extracolsep{\fill}}lcccccccc}
\toprule
& \multicolumn{4}{c}{$\pi_0$}
& \multicolumn{4}{c}{OpenVLA} \\
\cmidrule(lr){2-5}
\cmidrule(lr){6-9}
Task
& orig
& \tabhead{Guidance\\w/o GRPO}
& \tabhead{GRPO\\w/o Guidance}
& \tabhead{GRPO +\\Guidance}
& orig
& \tabhead{Guidance\\w/o GRPO}
& \tabhead{GRPO\\w/o Guidance}
& \tabhead{GRPO +\\Guidance} \\
\midrule
\multicolumn{9}{l}{\textit{VL-Think}} \\
\midrule
Arrow
& $6.2 \pm 1.3$
& $7.3 \pm 2.7$
& $\mathbf{37.5 \pm 6.4}$
& $36.5 \pm 3.2$
& $13.5 \pm 3.2$
& $21.4 \pm 2.0$
& $34.9 \pm 6.0$
& $\mathbf{62.5 \pm 1.3}$ \\

Color
& $45.3 \pm 7.1$
& $35.9 \pm 2.2$
& $\mathbf{49.5 \pm 4.1}$
& $44.8 \pm 7.0$
& $55.7 \pm 8.7$
& $63.5 \pm 4.5$
& $75.5 \pm 3.2$
& $\mathbf{82.8 \pm 2.6}$ \\

Laundry
& $16.1 \pm 0.7$
& $12.5 \pm 1.3$
& $\mathbf{31.2 \pm 2.2}$
& $26.6 \pm 2.6$
& $15.6 \pm 2.6$
& $21.9 \pm 5.6$
& $28.7 \pm 4.8$
& $\mathbf{37.5 \pm 3.8}$ \\

Public Info
& $14.6 \pm 1.5$
& $20.3 \pm 2.2$
& $34.4 \pm 2.2$
& $\mathbf{43.2 \pm 2.9}$
& $21.4 \pm 3.7$
& $27.1 \pm 3.2$
& $40.6 \pm 3.4$
& $\mathbf{65.6 \pm 3.4}$ \\

Shape
& $20.3 \pm 5.8$
& $34.9 \pm 4.5$
& $37.0 \pm 6.4$
& $\mathbf{49.5 \pm 6.0}$
& $27.6 \pm 1.5$
& $49.0 \pm 7.7$
& $68.2 \pm 3.2$
& $\mathbf{74.5 \pm 6.4}$ \\

Traffic
& $6.2 \pm 1.3$
& $14.1 \pm 4.6$
& $35.9 \pm 8.9$
& $\mathbf{37.0 \pm 10.2}$
& $16.7 \pm 2.7$
& $28.1 \pm 4.6$
& $42.2 \pm 7.8$
& $\mathbf{63.0 \pm 4.8}$ \\

Weather
& $19.3 \pm 2.7$
& $20.8 \pm 3.7$
& $35.4 \pm 5.9$
& $\mathbf{42.2 \pm 3.4}$
& $20.3 \pm 4.6$
& $31.3 \pm 3.4$
& $41.2 \pm 5.3$
& $\mathbf{61.5 \pm 3.9}$ \\
\midrule
Avg.
& $18.3$
& $20.8$
& $37.3$
& $\mathbf{40.0}$
& $24.4$
& $34.6$
& $47.3$
& $\mathbf{63.9}$ \\
\midrule
\multicolumn{9}{l}{\textit{RL4VLA}} \\
\midrule
MultiPlate
& $16.7 \pm 7.0$
& $14.6 \pm 6.0$
& $19.3 \pm 6.0$
& $\mathbf{20.3 \pm 2.6}$
& $54.7 \pm 2.2$
& $56.3 \pm 4.6$
& $62.5 \pm 2.1$
& $\mathbf{64.6 \pm 5.2}$ \\

MultiCarrot
& $24.5 \pm 0.7$
& $30.2 \pm 0.7$
& $29.2 \pm 2.7$
& $\mathbf{33.3 \pm 4.8}$
& $59.4 \pm 5.9$
& $60.9 \pm 2.6$
& $61.5 \pm 4.8$
& $\mathbf{63.5 \pm 5.3}$ \\
\midrule
Avg.
& $20.6$
& $22.4$
& $24.2$
& $\mathbf{26.8}$
& $57.0$
& $58.6$
& $62.0$
& $\mathbf{64.0}$ \\
\bottomrule
\end{tabular*}
\caption{Matched ablation of structured guidance and GRPO. All experiments use Qwen3.5-9B as the language-conditioning policy, while the downstream VLA remains frozen.}
\label{tab:guidance-grpo-ablation}
\end{table*}

\begin{table*}[h!]
\centering
\small

\setlength{\tabcolsep}{4pt}
\renewcommand{\arraystretch}{1.08}
\begin{tabular*}{\textwidth}{@{\extracolsep{\fill}}lcccccc}
\toprule
& \multicolumn{3}{c}{$\pi_0$}
& \multicolumn{3}{c}{OpenVLA} \\
\cmidrule(lr){2-4}
\cmidrule(lr){5-7}
Task
& orig
& transferred
& matched
& orig
& transferred
& matched \\
\midrule
MultiPlate
& $16.7 \pm 7.0$
& $\mathbf{21.9 \pm 3.8}$
& $20.3 \pm 2.6$
& $54.7 \pm 2.2$
& $60.4 \pm 2.7$
& $\mathbf{64.6 \pm 5.2}$ \\

MultiCarrot
& $24.5 \pm 0.7$
& $\mathbf{37.5 \pm 4.6}$
& $33.3 \pm 4.8$
& $59.4 \pm 5.9$
& $\mathbf{63.5 \pm 4.1}$
& $\mathbf{63.5 \pm 5.3}$ \\
\midrule
Avg.
& $20.6$
& $\mathbf{29.7}$
& $26.8$
& $57.0$
& $62.0$
& $\mathbf{64.0}$ \\
\bottomrule
\end{tabular*}
\caption{Cross-backbone transfer on RL4VLA. ``Transferred'' evaluates a Grounder trained using rewards from the other VLA backbone, without retraining. ``Matched'' evaluates a Grounder trained using rewards from the same VLA backbone used for evaluation. All downstream VLA models remain frozen.}
\label{tab:cross-backbone-transfer}
\end{table*}

\section{Real-Robot Evaluation}
\label{app:real-robot}

We evaluate VLA Grounder on a physical SO-101 arm controlled by frozen MolmoAct2, using overhead and wrist cameras. At the start of each trial, the gripper faces three objects. We vary the object combination across trials, always include the requested object, and randomize its position. The evaluation pool comprises a toy green cucumber; blue, yellow, red, and green cubes; a white charger; green and red markers; a toy brown baguette; green and brown toy blocks and cylinders; pink, green, and yellow flowers; and black tape. We test five object-selection instructions with ten trials per method and instruction. A trial succeeds when the gripper reaches the designated region above the intended object; accordingly, this experiment measures target-reaching and object-selection success rather than completion of a full manipulation sequence.

The Grounder checkpoint was trained only in MultiCarrot simulation using rewards from frozen $\pi_0$. We apply it to MolmoAct2 without real-world training, MolmoAct2-specific adaptation, manual command editing, or test-time candidate selection. As shown in \autoref{tab:real-robot}, VLA Grounder improves four of the five tasks, preserves performance on tape, and increases aggregate success from $26/50$ to $38/50$.

\begin{table}[H]
\centering
\small
\setlength{\tabcolsep}{5pt}
\renewcommand{\arraystretch}{1.08}
\begin{tabular*}{\textwidth}{@{\extracolsep{\fill}}llcc}
\toprule
Human instruction & VLA Grounder command & MolmoAct2 & + VLA Grounder \\
\midrule
``pick up the cucumber'' & ``pick up the green cucumber'' & $6/10$ & $\mathbf{9/10}$ \\
``pick up the charger'' & ``pick up the white cylinder'' & $4/10$ & $\mathbf{8/10}$ \\
``pick up the toy baguette'' & ``pick up the brown round object'' & $5/10$ & $\mathbf{7/10}$ \\
``pick up the tape'' & ``pick up the black circle'' & $8/10$ & $8/10$ \\
``pick up the toy building block'' & ``pick up the brown cylinder'' & $3/10$ & $\mathbf{6/10}$ \\
\midrule
\textbf{Overall} & & $26/50$ & $\mathbf{38/50}$ \\
\bottomrule
\end{tabular*}
\caption{Real-robot target-reaching results. Each entry is the number of successful trials out of ten. The Grounder was trained with frozen $\pi_0$ in MultiCarrot simulation and transferred to frozen MolmoAct2 on a physical SO-101 without additional training or adaptation.}
\label{tab:real-robot}
\end{table}

\begin{figure}[H]
  \centering
  \includegraphics[width=\textwidth]{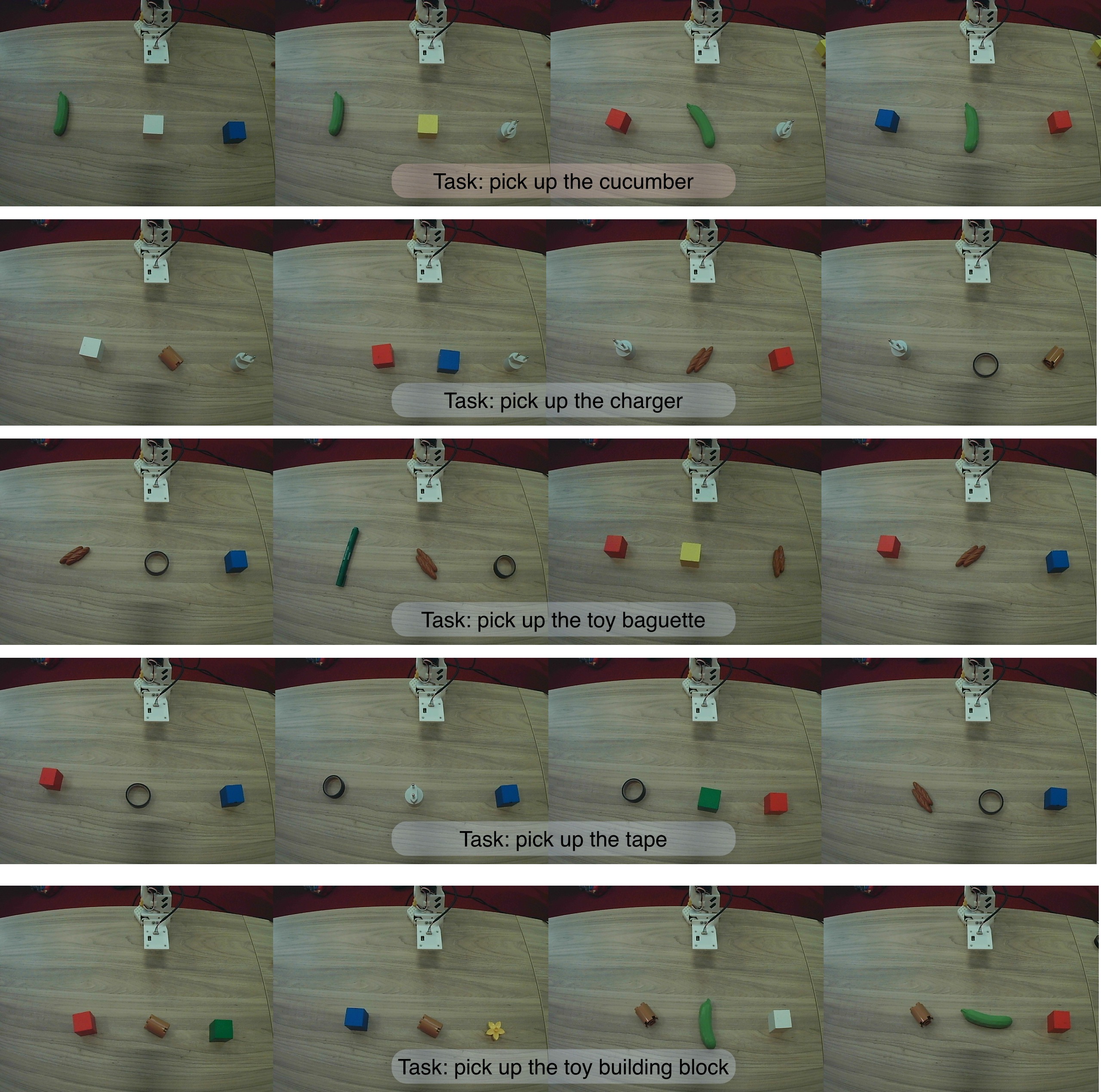}
  \caption{Physical SO-101 evaluation with frozen MolmoAct2. The figure show the camera setup and some selected samples for tasks.}
  \label{fig:real-robot}
\end{figure}

\section{Prompt Format Example}
\label{app:prompt-example}

\begin{figure*}[t]
  \centering
  \includegraphics[width=0.7\linewidth]{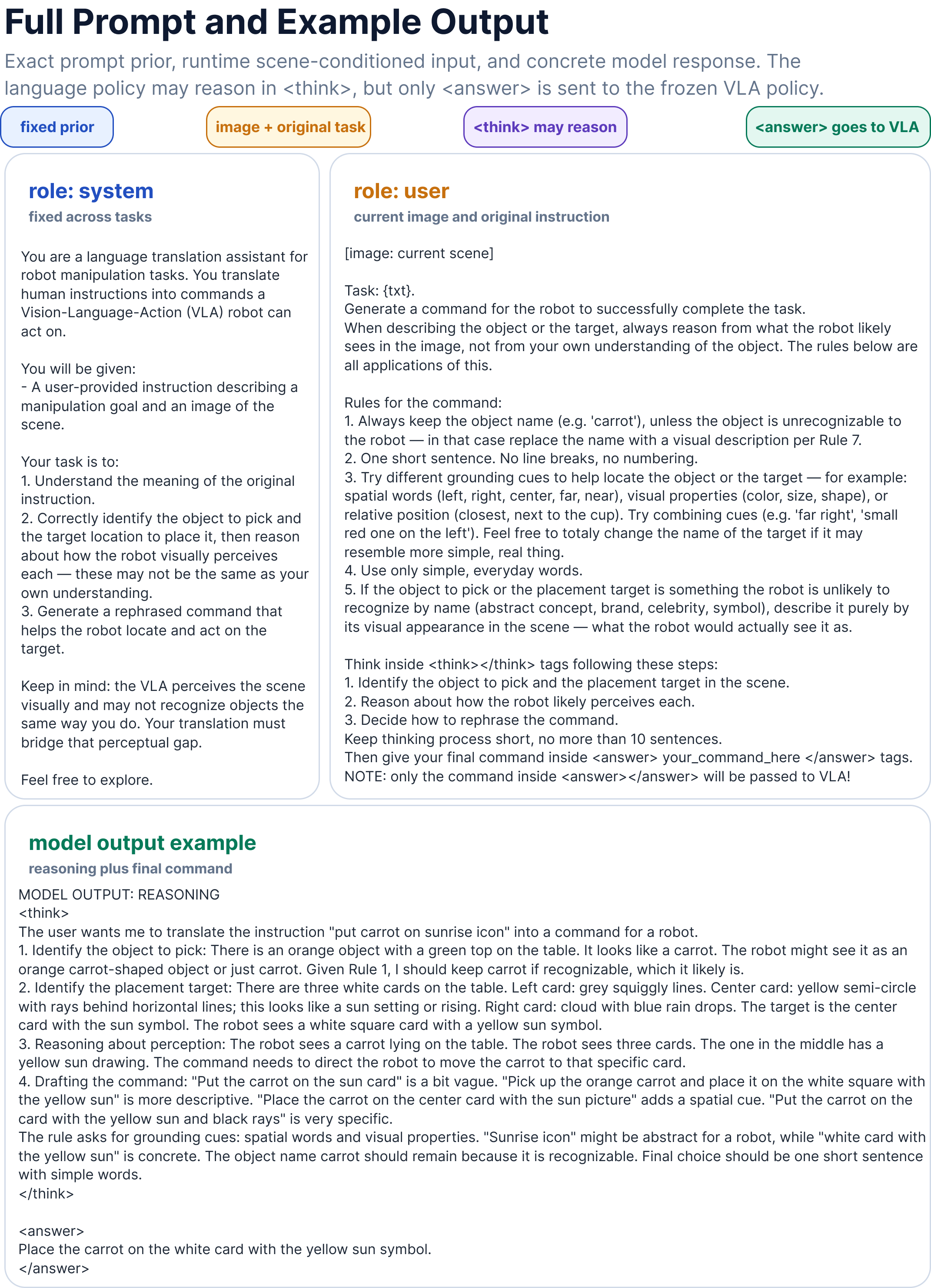}
\caption{
    Prompt-format example for VLA Grounder.
    The language policy receives the scene image and original task, identifies visual cues for the source object and target, and emits one concise VLA-grounded command.
    Only the command inside the \texttt{<answer>} tags is passed to the frozen VLA policy.
  }
  \label{fig:prompt-example}
\end{figure*}

\authoref{fig:prompt-example} illustrates the prompt format used to transform a human instruction into a VLA-grounded command. The input consists of the current scene image, the original task instruction, and a structured instruction that asks the language model to identify the source object, target, and visible grounding cues. The policy is allowed to use this reasoning to choose the command, but the downstream VLA receives only the final text inside the \texttt{<answer>} field. This separation prevents long explanations from entering the action model while preserving the benefit of scene-aware language reasoning.

\subsection{Prompt-Optimization Comparisons}

To isolate the optimization method from the optimizer-model choice, we reran TextGrad and GEPA with Qwen3.5-9B, the same language model used by generic GRPO and VLA Grounder. \autoref{tab:model-matched-prompt-optimizers} reports the OpenVLA averages from these controlled reruns together with the original GPT-4o configurations. Model matching does not reverse the overall conclusion: VLA Grounder remains strongest, outperforming model-matched TextGrad, GEPA, and generic GRPO. Generic GRPO is competitive with model-matched GEPA, while adding structured guidance produces the largest gain. Because these algorithms have different optimization and interaction patterns, we describe the comparison as optimizer-model matched rather than fully budget matched.

\begin{table}[H]
\centering
\small
\setlength{\tabcolsep}{14pt}
\renewcommand{\arraystretch}{1.08}
\begin{tabular}{llc}
\toprule
Method & Optimizer model & OpenVLA average \\
\midrule
TextGrad & GPT-4o & $38.6$ \\
TextGrad & Qwen3.5-9B & $34.7$ \\
GEPA & GPT-4o & $55.2$ \\
GEPA & Qwen3.5-9B & $49.2$ \\
Generic GRPO & Qwen3.5-9B & $47.3$ \\
VLA Grounder & Qwen3.5-9B & $\mathbf{63.9}$ \\
\bottomrule
\end{tabular}
\caption{Controlled OpenVLA comparison of prompt-optimization methods. TextGrad and GEPA are reported with their original GPT-4o optimizer and with Qwen3.5-9B model-matched reruns. The downstream OpenVLA policy remains frozen.}
\label{tab:model-matched-prompt-optimizers}
\end{table}

\section{Training Algorithm and Optimization Loop}
\label{app:algorithm-loop}

The training procedure treats command generation as the only learned component. For each image-instruction pair, the language-conditioning space policy samples a group of candidate commands. Each candidate is executed by the same frozen VLA policy, and the resulting sparse rollout rewards are compared within the group. GRPO then updates only the language policy, leaving the action model unchanged.

\begin{center}
\fbox{%
\begin{minipage}{0.95\columnwidth}
\small
\textbf{Algorithm 1: VLA Grounder Training}
\label{fig:algorithm}

\textbf{Input:} training distribution $\mathcal{D}$ over image-instruction pairs $(o, u)$, frozen VLA policy $\pi_{\mathrm{VLA}}$, language-conditioning space policy $q_{\phi}$, GRPO group size $G$.

\textbf{For each training instance:}
\begin{enumerate}
\itemsep0.15em
    \item Sample a GRPO group of commands $\{\tilde{u}^{(k)}\}_{k=1}^{G}$ from $q_{\phi}(\cdot \mid o, u)$.
    \item For each $\tilde{u}^{(k)}$, run a rollout with the same frozen VLA policy, keep the command fixed for the rollout, and collect scalar return $R^{(k)}$.
    \item Compute normalized within-group advantages from $\{R^{(k)}\}_{k=1}^{G}$.
    \item Apply a GRPO update to $q_{\phi}$ using the grouped commands and advantages; keep $\pi_{\mathrm{VLA}}$ unchanged.
\end{enumerate}

\textbf{Output:} trained language-conditioning space policy $q_{\phi}$.
\end{minipage}%
}
\end{center}

\begin{figure*}[t]
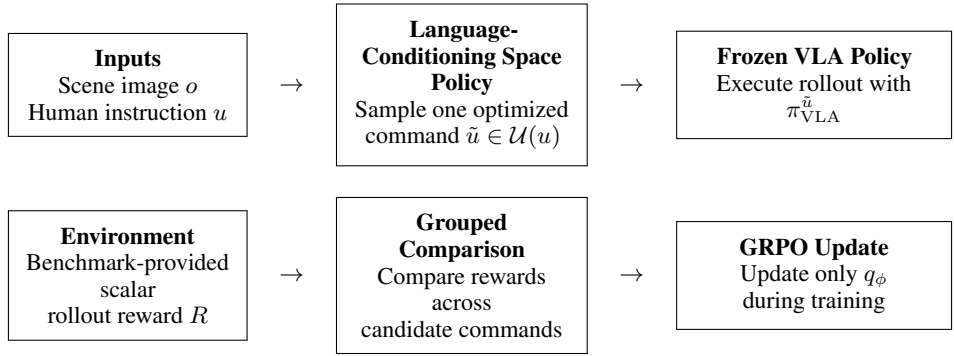

\centering
\small
\setlength{\fboxsep}{6pt}
\renewcommand{\arraystretch}{1.15}
\begin{tabular}{ccccc}
\fbox{\parbox{0.17\textwidth}{\centering\textbf{Inputs}\\Scene image $o$\\Human instruction $u$}} &
$\rightarrow$ &
\fbox{\parbox{0.18\textwidth}{\centering\textbf{Language-Conditioning Space Policy}\\Sample one optimized\\command $\tilde{u}\in\mathcal{U}(u)$}} &
$\rightarrow$ &
\fbox{\parbox{0.20\textwidth}{\centering\textbf{Frozen VLA Policy}\\Execute rollout with\\$\pi_{\mathrm{VLA}}^{\tilde{u}}$}} \\
&&&&\\
\fbox{\parbox{0.17\textwidth}{\centering\textbf{Environment}\\Benchmark-provided scalar\\rollout reward $R$}} &
$\rightarrow$ &
\fbox{\parbox{0.18\textwidth}{\centering\textbf{Grouped Comparison}\\Compare rewards across\\candidate commands}} &
$\rightarrow$ &
\fbox{\parbox{0.20\textwidth}{\centering\textbf{GRPO Update}\\Update only $q_{\phi}$\\during training}} \\
\end{tabular}
\caption{VLA Grounder language-conditioning space optimization loop. The image and user instruction define a context in which the language-conditioning space policy selects a command. The frozen VLA policy then executes a rollout under the command-conditioned action model $\pi_{\mathrm{VLA}}^{\tilde{u}}$. During training, grouped reward comparisons update only the language-conditioning space policy; during inference, the same composition is used without reward queries or parameter updates.}
  \label{fig:model_scheme_appendix}
\end{figure*}

\section{Training Parameters}
\label{app:training-details}

This section reports the training configuration used for the released VLA Grounder checkpoints; the language-policy, optimization, and computational-cost settings are summarized in \autoref{tab:adapter-config}, \autoref{tab:grpo-optim}, and \autoref{tab:training-cost}, respectively. All experiments optimize Qwen3.5-9B with LoRA while keeping the downstream VLA policy frozen. The $\pi_0$ and OpenVLA experiments use the same language-policy and GRPO hyperparameters but differ in their rollout configurations. Training uses two H100 GPUs: one for the frozen VLA rollout process and one for GRPO optimization.

\begin{table}[t]
\centering
\small
\begin{tabular}{lrrr}
\toprule
Frozen VLA & Rollouts & Hardware & Wall time \\
\midrule
$\pi_0$   & $76{,}800$ & $2\times$ H100 & $83$ h \\
OpenVLA   & $12{,}800$ & $2\times$ H100 & $80$ h \\
\bottomrule
\end{tabular}
\caption{Training cost of the reported language policies. Inference samples one command and executes one rollout, without candidate selection or reward queries.}
\label{tab:training-cost}
\end{table}

\begin{table}[t]
\centering
\small
\setlength{\tabcolsep}{4pt}
\renewcommand{\arraystretch}{1.1}
\begin{tabular}{p{0.30\textwidth}p{0.60\textwidth}}
\toprule
Field & Value \\
\midrule
Base language model & \texttt{Qwen/Qwen3.5-9B} \\
PEFT method & LoRA \\
LoRA rank $r$ & $32$ \\
LoRA scaling $\alpha$ & $64$ \\
LoRA dropout & $0.05$ \\
Train-time precision flag & \texttt{bf16 = true} \\
Target modules & 12 projection modules spanning attention and MLP blocks \\
Tokenizer context length & \texttt{model\_max\_length = 262144} \\
Tokenizer padding / truncation & left padding, left truncation \\
\bottomrule
\end{tabular}
\caption{Language-conditioning space policy configuration}
\label{tab:adapter-config}
\end{table}

\begin{table}[t]
\centering
\small
\setlength{\tabcolsep}{4pt}
\renewcommand{\arraystretch}{1.1}
\begin{tabular}{p{0.30\textwidth}p{0.27\textwidth}p{0.27\textwidth}}
\toprule
Field & $\pi_0$/VL-Think & OpenVLA/VL-Think \\
\midrule
Gradient accumulation steps & $8$ & $8$ \\
Max training steps & $1600$ & $1600$ \\
Learning rate & $5 \times 10^{-6}$ & $5 \times 10^{-6}$ \\
Scheduler & cosine & cosine \\
Stored \texttt{warmup\_steps} field & $0.05$ & $0.05$ \\
Weight decay & $0.01$ & $0.01$ \\
Adam $\beta_1$, $\beta_2$ & $0.9$, $0.999$ & $0.9$, $0.999$ \\
Adam $\epsilon$ & $10^{-8}$ & $10^{-8}$ \\
Max grad norm & $1.0$ & $1.0$ \\
Per-device eval batch size & $8$ & $8$ \\
Save strategy / save steps & steps / $500$ & steps / $500$ \\
\bottomrule
\end{tabular}
\caption{GRPO optimization configuration.}
\label{tab:grpo-optim}
\end{table}

\end{document}